\documentclass[runningheads]{llncs}

\usepackage{graphicx}
\usepackage{amsmath,amssymb} 
\usepackage{color}
\usepackage{tumabbrev}
\usepackage{glossaries}
\newacronym[]{gls:RS}{RS}{remote sensing}
\newacronym[]{gls:GSD}{GSD}{ground sampling distance}
\newacronym[]{gls:CNN}{CNN}{convolutional neural network}
\newacronym[]{gls:CRF}{CRF}{conditional random field}
\newacronym[]{gls:SGD}{SGD}{Stochastic Gradient Descent}
\newacronym[]{gls:FCNN}{FCNN}{fully convolutional neural network}
\newacronym[]{gls:DCNN}{DCNN}{deep convolutional neural network}
\newacronym[]{gls:ACNN}{ACNN}{atrous convolutional neural network}
\newacronym[longplural={oriented bounding boxes}]{gls:OBB}{OBB}{oriented bounding box}
\newacronym[longplural={horizontal bounding boxes}]{gls:HBB}{HBB}{horizontal bounding box}
\newacronym[]{gls:UAV}{UAV}{unmanned aerial vehicles}
\newacronym[]{gls:miou}{mIoU}{mean intersection over union}
\newacronym[]{gls:GFLOP}{GFLOP}{giga floating point operation}
\newacronym[]{gls:RCNN}{RCNN}{Region-based convolutional neural network}
\newacronym[]{gls:SSD}{SSD}{multi-box single shot detector}
\newacronym[]{gls:DAB}{DAB}{domain adapter block}
\newacronym[]{gls:NMS}{NMS}{non-maximum suppression}
\newacronym[]{gls:FPS}{FPS}{frame per second}
\newglossaryentry{gls:FastRCNN}{name={Fast-\gls{gls:RCNN}}, description={}}
\newglossaryentry{gls:FasterRCNN}{name={Faster-\gls{gls:RCNN}}, description={}}
\newglossaryentry{gls:MaskRCNN}{name={Mask-\gls{gls:RCNN}}, description={}}
\usepackage{hyperref}
\usepackage[noabbrev, capitalize]{cleveref}
\usepackage[sectionbib,numbers]{natbib}

\usepackage{adjustbox}
\hypersetup{%
colorlinks=true,
urlbordercolor={1 1 1},
linkcolor={red},
raiselinks=false
}
\usepackage{color,soul}
\usepackage{booktabs}
\usepackage{lipsum}
\usepackage{multirow}
\usepackage{multicol}
\usepackage{subcaption}
\usepackage{siunitx}
\DeclareSIUnit\px{px}

\usepackage{xspace}

\usepackage{etoolbox}
\patchcmd{\thebibliography}{\chapter*}{\section*}{}{}
\begin{document}\sloppy
\title{ShuffleDet: Real-Time Vehicle Detection Network in On-board Embedded UAV Imagery} 

\titlerunning{ShuffleDet}
%
\author{Seyed~Majid~Azimi\inst{1,2}\orcidID{0000-0002-6084-2272}}
%
\authorrunning{Seyed Majid Azimi}
%

\institute{German Aerospace Center (DLR), Remote Sensing Technology Institute,  \\We{\ss}ling, Germany \\
\and
Technical University of Munich, Chair of Remote Sensing, 
\\Munich, Germany\\
\email{seyedmajid.azimi@\{dlr,tum\}.de}}
\maketitle              
\begin{abstract}
On-board real-time vehicle detection is of great significance for UAVs and other embedded mobile platforms.
We propose a computationally inexpensive detection network for vehicle detection in UAV imagery which we call ShuffleDet. In order to enhance the speed-wise performance, we construct our method primarily using channel shuffling and grouped convolutions. We apply inception modules and deformable modules to consider the size and geometric shape of the vehicles. ShuffleDet is evaluated on CARPK and PUCPR+ datasets and compared against the state-of-the-art real-time object detection networks. ShuffleDet achieves 3.8 GFLOPs while it provides competitive performance on test sets of both datasets. We show that our algorithm achieves real-time performance by running at the speed of 14 frames per second on NVIDIA Jetson TX2 showing high potential for this method for real-time processing in UAVs.

\end{abstract}
\section{Introduction}On-board real-time processing of data through embedded systems plays a crucial role in applying the images acquired from the portable platforms (e.g., \glspl{gls:UAV}) to the applications requiring instant responses such as search and rescue missions, urban management, traffic monitoring, and parking lot utilization.

Methods based on \glspl{gls:CNN}, for example, FPN~\cite{fpn}, FasterRCNN~\cite{fasterrcnnNIPS2015}, R-FCN~\cite{NIPS2016_6465}, \glspl{gls:SSD}~\cite{DBLP:conf/eccv/LiuAESRFB16}, and Yolov2~\cite{redmon2017yolo9000}, have shown promising results in many object detection tasks. Despite their high detection precision, these methods are computationally demanding and their models are usually bulky due to the deep backbone networks being used.
Employing \glspl{gls:CNN} for the on-board real-time applications requires developing time and computation efficient methods due to the limited processing resources available on-board.
A number of networks have been developed recently such as GoogleNet~\cite{inception}, Xception~\cite{xception}, ResNeXt~\cite{resnext}, MobileNet~\cite{mobilenetv1}, PeleeNet~\cite{pelee}, SqueezeNet~\cite{squeeznet}, and ShuffleNet~\cite{ZhaZho17ShuffleNet} which have less complex structures as compared to the other \glspl{gls:CNN} while providing comparable or even superior results.
For the real-time object detection applications (\eg vehicle detection), there are a few recent works proposing the methods such as MobileNet~\cite{mobilenetv1} with SSD~\cite{8099834}, PVANET~\cite{pvanet}, and Tiny-Yolo~\cite{redmon2017yolo9000}. They have shown computational efficiency to be deployed in mobile platforms.

Zhang et al.~\cite{ZhaZho17ShuffleNet} employed ShuffleNet as the backbone network, which uses point-wise grouped convolutions and channel shuffle to greatly reduce the computations while maintaining the accuracy.
The authors reported a better performance compared with MobileNet using Faster-RCNN detection approach.
Kim et al.~\cite{pvanet} developed PVANET by concatenating $3\times3$ conv layer with its negation as a building block for the initial feature extraction stage.
Recently, Wang et al.~\cite{pelee} proposed PeleeNet that uses a combination of parallel multi-size kernel convolutions as a 2-way dense layer and a similar module to the Squeeze module.
They additionally applied a residual block after feature extraction stage to improve the accuracy using the SSD~\cite{DBLP:conf/eccv/LiuAESRFB16} approach. The authors reported more accurate results compared to MobileNet and ShuffleNet on the Pascal VOC dataset despite the smaller model size and computation cost of  PeleeNet.
Redmon and Farhadi~\cite{redmon2017yolo9000} proposed Yolov2, a fast object detection method, but yet with high accuracy.
However, their method is still computationally heavy for real-time processing on an embedded platform.
Tiny Yolov2 as the smaller version of Yolov2, although it is faster, but it lacks high-level extraction capability which results in poor performance.
In the work of Huang et al.~\cite{8099834}, they showed the SSD detection approach together with SqueezeNet and MobileNet as the backbone networks.
Although SSD with SqueezeNet backbone results in a smaller model than MobileNet, its results are less accurate and its computation is slightly more expensive.
In general, replacing the backbone network with SqueezeNet, MobileNet, or any other efficient network - though enhancing computational efficiency - can degrade the accuracy if no further modification is performed.

In this paper, we propose ShuffleDet, a real-time vehicle detection approach to be used on-board by mobile platforms such as \glspl{gls:UAV}. ShuffleDet network is composed of ShuffleNet and a modified variant of SSD based on channel shuffling and grouped convolution.
We design a unit to appropriately transfer the pretrained parameters of the pretrained model on terrestrial imagery to aerial imagery domain.
We call this unit ~\gls{gls:DAB} which includes deformable convolutions~\cite{dai17dcn} and Inception-ResNetv2 units~\cite{inceptionresnetv2}.
To the best of our knowledge, group convolution and channel shuffling have not been used before in real-time vehicle detection based on \gls{gls:UAV} imagery.
ShuffleDet runs at 14 frames per second (FPS) on NVIDIA Jetson TX2 while having the computational complexity of 3.8 \glspl{gls:GFLOP}.
Experimental results on the CARPK and PUCPR+ datasets~\cite{HsiehLH17} demonstrates that ShuffleDet achieves a good trade-off between accuracy and speed for mobile platforms while it is comparably computation and time efficient.

\section{Method}
	In this section, a detailed description of the network architecture is presented.
	We use ShuffleNet~\cite{ZhaZho17ShuffleNet} which is designed for object recognition to extract high-level features as our backend network.

	\begin{figure*}[h]
		\centering
		\includegraphics[width=\textwidth]{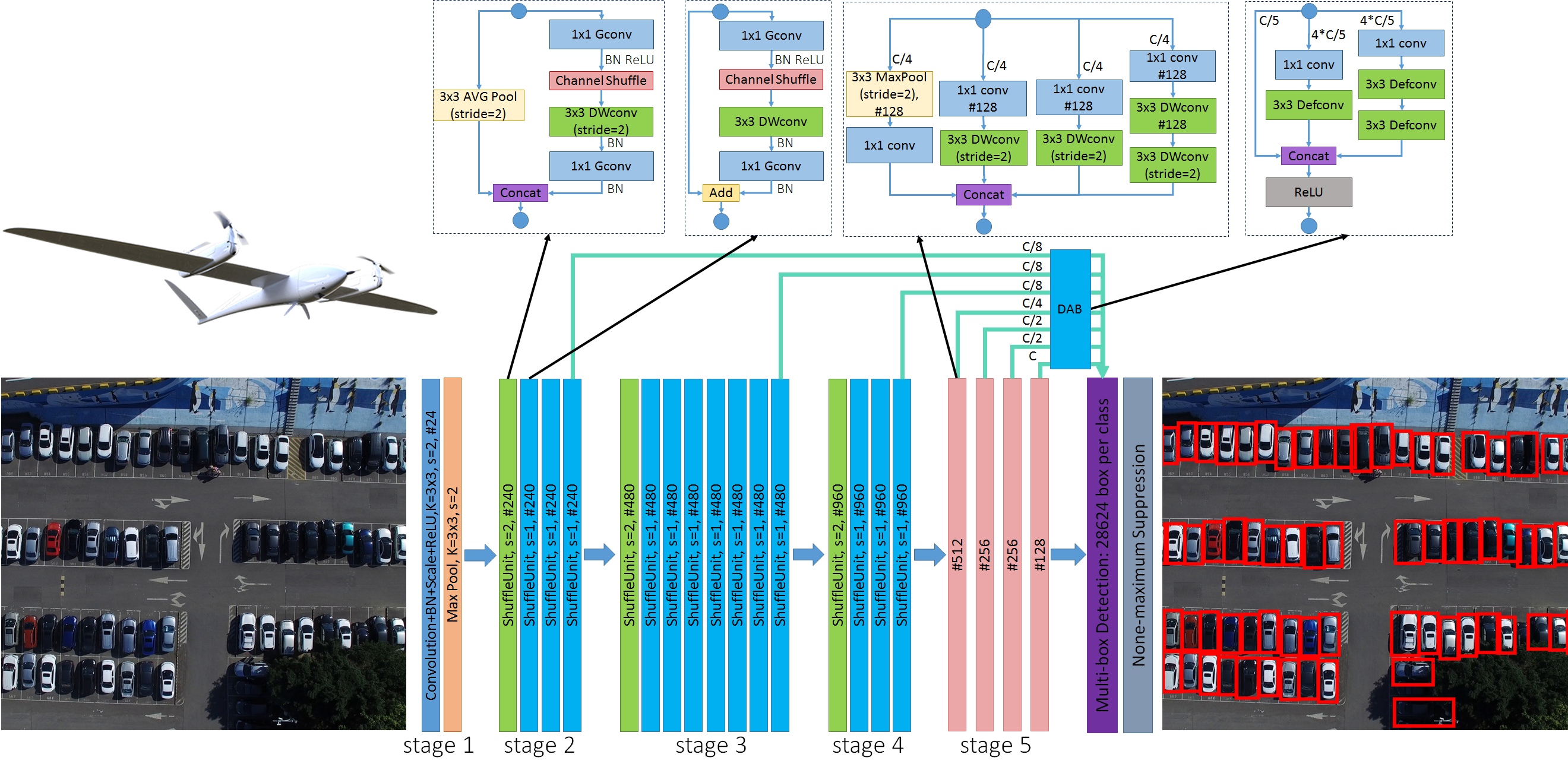}
		\caption{Illustration of ShuffleDet architecture.
        The backbone network is ShuffleNet.
        Modified inception layers are applied as extra layers.
        $C$ stands for channel.
        \gls{gls:DAB} unit is deployed to adapt to the new domain of \gls{gls:UAV} imagery using a residual block containing deformable convolution layers\protect\footnotemark.}
        \label{fig:overflow}
	\end{figure*}
    \footnotetext{UAV photo is from \url{https://www.quantum-systems.com/tron}}
	ShuffleNet\cite{ZhaZho17ShuffleNet} shows that by utilizing grouped or depth-wise separable convolutions, one can reduce the computational demand, while still boosting the performance through a decent representation ability.
    A major bottleneck can arise by replacing $1\times1$ convolution layers with stacked grouped convolutions which can degrade the accuracy of the network.
    This is due to the fact that a limited portion of input channels are utilized by the output channels.
    In order to solve this issue channel shuffling was proposed in \cite{ZhaZho17ShuffleNet} which we also use inside ShuffleDet architecture. Figure~\ref{fig:overflow} illustrates the network architecture of ShuffleDet.
    In stage 1, a $3\times3$ convolutional layer is applied to the input image with a stride of 2 which downsamples the input by a factor of 2.
    This layer is followed by a maxpooling layer with a stride of 2 and kernel of $3\times3$. This maxpooling operation destroys half of the input information.
    This is critical as vehicles in our case are small objects~\cite{azimiACCV,azimi2018ISPRS,azimi2018advanced, azimi2018aerial}. Having said that without this operation, computation cost will be multiplied.
    Therefore, we keep the maxpooling layer and we try to enhance the performance especially via \glspl{gls:DAB} units which will be discussed later.
    After the maxpooling three stages containing multiple units from ShuffleNet are performed.
    Stage 2 and 4 contain 3 ShuffleNet units while stage 3 in the middle is composed of 7 units.
    The whole stage 1 to 4 leads to 32x down-sampling factor. ShuffleUnit illustrated in Figure~\ref{fig:overflow} acts as residual bottleneck unit.
	Using stride 2 in ShuffleUnit, an average pooling is applied to the primary branch parallel with depthwise convolution with a stride 2 in the residual branch.
    To ensure that all of the input channels are connected to the output channels, channel shuffling is performed before the depthwise convolution.
    A $1\times1$ grouped convolutions are applied before the channel shuffling as a bottleneck in order to reduce the number of feature maps in the output for the efficient computation.
    It has been shown~\cite{ZhaZho17ShuffleNet,quantization} that the group convolutions also improve the accuracy.
    The second grouped convolution brings back the number of feature maps or channel to the number of input channels for a more accurate representation capability.
    Using a stride of 2, the features of average pooling and second grouped convolution is concatenated while having a stride of 1, maxpooling is omitted and depth-wise convolution is performed.
    Moreover, the outputs are summed up instead of using concatenation.
	Figure \ref{fig:overflow} shows the detailed structure of ShuffleNet units with and without stride of 2.
	
	Stage 1, 2, 3 and stage 4 are employed to enhance the heat map resolution as input intermediate layers.
	In the detection module, we primarily inspire from SSD approach.
	In order to enrich the extracted features from the intermediate layers, we perform extra feature layers in stage 5.
	In our case, the output from stage 4 is passed through stage 5 as illustrated in Figure~\ref*{fig:overflow}
	This is compatible with using multi-box strategy explained in the SSD method.
	In total, we extract 7 feature maps of different sizes from the backbone network.
	
    To enhance the performance, instead of employing a conventional convolution layer similar to SSD method for each extra layer, we use a modified module of Reduction-B from Inception-ResNet-v2~\cite{inceptionresnetv2} in stage 5.
	Unlike ResNet and VGG, inception modules have not been explored enough in object detection task due to their higher computation cost.
	We stack 4 modified inception modules as stage 5 for feature map extraction at different levels.
	Unlike original Inception-ResNet-v2 work, we add $1\times1$ conv layers after maxpooling and concatenation layer.
	The maxpooling layer reduces spatial-resolution and dimension as a bottleneck.
	$1\times1$ convolution in return expands the dimension to insert further non-linearity to the network resulting in a better performance. 
	The same philosophy was used in the latter $1\times1$ conv layer. 
	Applying the inception module adds more computational cost to the network.
	To compensate its load, we replace $3\times3$ convolution layers with $3\times3$ depthwise convolutions.
	Depth-wise convolution improves the performance slightly, yet it has $\dfrac{1}{N} + \dfrac{1}{k^2}$ times less computation cost compared with regular conv layers.
	$N$ is the number of output channels and $k$ is the kernel size.
	Furthermore, we divide the input channels equally among the branches.
	The output number of channels for each layer is an equally-divided concatenation of output channels from each branch.
	These modifications keep the model size as well as computational complexity small.
	We observe using this modified inception modules enhances the performance.
	We conjecture unlike the original SSD which uses $1\times1$ and $3\times3$ conv layers in series as extra layers, multi-size kernels parallel in inception modules capture features in different sizes simultaneously e.g. $1\times1$ kernels to detect small vehicles and $3\times3$ kernels for bigger ones which could be the reason for this enhancement.
	This shows by widening the network and augmenting the cardinality, we can achieve better results.
	This comes only with a marginal increase in computational complexity.
	Moreover, by using multi-size kernels, one does not need to worry which kernel size is more appropriate.

	In order to regress bounding boxes and predict object classes from extra layers as illustrated in \cref{fig:overflow}, the base-line SSD processes each feature map by only a single $3\times3$ convolution layer followed by {\fontfamily{qcr}\selectfont permute} and {\fontfamily{qcr}\selectfont flatten} layers in multi-box detection layer.
	This includes feature maps only from one of the high-resolution layers.
	This leads to a weak performance in detecting small-scale vehicles.
	The feature maps from higher-resolution layers e.g. in our case stage 2 and 3 are responsible to detect small-scale vehicles.
	Stage 1 is ignored due to its high computational complexity.
	Those corresponding feature maps are semantically weak and not deep enough to be capable of detecting small-scale vehicles.
	ResNet and VGG19 works denote that employing deeper features enhances the object recognition accuracy.
	However, those backbone networks are computationally heavy to be deployed on on-board processors in UAVs which work under strict power constraints.
	As an alternative, we propose using a residual module which we call DAB as shown in Figure~\ref{fig:overflow}.
	Combination of $1\times1$ convention and $3\times3$ deformable convolution operations enrich the features further, but still introducing low computation burden.
	We choose a portion of input channels to keep the computation cost low.
	${1/8, 1/8, 1/8, 1/4, 1/2, 1/2, 1}$ are used as the portion of input channels of output layers from stage 2 to the last extra layer and inside DAB unit we assign ${1/5, 4/5, 4/5}$ portion of input channels to each branch as illustrated in Figure~\ref{fig:overflow}.
	The output channels remain similar to the original SSD.
	The only difference is the introduced extra multi-box feature map from stage 2.
	SSD calculates the number of default boxes per class by $W\times H \times B$ in which $W$ and $H$ are input width and height and $B$ is from the set of ${4, 6, 6, 4, 4}$ for each feature map.
	We choose $B=4$ for the stage 2 leading to 28642 boxes per class.
	
	In aerial imagery, vehicles appear to be very small and almost always in rectangle geometric shape.
	On the other hand, the pre-trained ShuffleNet has been trained on ground imagery while our images are in another domain of aerial imagery.
	Therefore pre-trained weights should be adapted to the new domain.
	We use deformable convolution as introduced in \cite{dai17dcn} to take into account the new domain and the geometric properties of the vehicles.
	Deformable convolution adds an offset to the conventional conv layer in order to learn from the geometric shape of the objects.
	They are not limited to a fix kernel size and offset is learned during training by adding only an inexpensive conv layer to compute the offset field.
	Deformable conv layer shows considerable improvement in case of using images acquired from low-flying UAVs. 
	However, the impact is less by using images from high-altitude platforms such as helicopter or airplanes.
	According to \cite{dai17dcn} the computation cost of deformable convolutions is negligible. 
	Finally, we apply {\fontfamily{qcr}\selectfont ReLU} layer to element-wise added features in the DAB to add more non-linearity.
	In general, naive implementation of ShuffleNet with SSD has 2.94 GFLOPs while ShuffleDet has 3.8 GFLOPs.
	Despite an increase in the computation cost, ShuffleDet has considerable higher accuracy.
	As vehicles appear to be small objects in UAV images, we choose default prior boxes with smaller scales similar to \cite{azimiACCV}.
	Eventually, \gls{gls:NMS} is employed to suppress irrelevant detection boxes.
	It is worth mentioning that during training hard negative mining is employed with the ratio of $3:1$ between negative and positive samples.
	This leads to more stable and faster training.
	We also apply batch normalization after each module in DAB as well as extra feature layers.

	\section{Experiments and Discussion}
	In this section, we provide ablation evaluation of our proposed approach and compare it to the state-of-the-art CNN-based vehicle detection methods. The experiments were conducted on the CARPK and PUCPR+ datasets~\cite{HsiehLH17}, which contain 1573 and 125 images of $1280\times720$ pixels, respectively. The vehicles in the images are annotated by horizontal bounding boxes.
	To have a fair comparison with different baseline methods, we follow the same strategy as theirs for splitting the datasets into training and testing sets.
Moreover, we train ShuffleNet as the backbone network on the ImageNet-2012~\cite{imagenet} dataset achieving similar performance compared to the original ShuffleNet work. The results are compared to the benchmark using MAE and RMSE, similar to the baseline~\cite{HsiehLH17}.
	In addition, we use data augmentation in a similar way to the original work on SSD. 
	\subsection{Experimental Setup}
	We use Caffe to implement our proposed algorithm. It is trained using Nvidia Titan XP GPU and evaluated on NVIDIA Jetson TX2 as an embedded edge device.
	For the optimization, we use stochastic gradient descent with the base learning rate of 0.001, gamma 0.1, momentum 0.9 to train the network for 120k iterations. The learning rate is reduced after 80k and 100k by a factor of 10.
	Moreover, the images are resized to $512\times512$ pixels along with their annotations.
	Additionally, we initialize the first four layers with our pre-trained ShuffleNet weights and the rest with Gaussian noise.
	For the grouped convolutions, we set the number of groups to 3 throughout the experiments. Furthermore, \gls{gls:NMS} of 0.3 and confidence score threshold of 0.5 are considered.
	\subsection{Ablation Evaluation}
	In this section, we present an ablation study on the effect of the submodules in our approach.
	Table~\ref{tab:1} shows the impact of the modified inception module compared to the original baseline.
	According to the results, introducing the first modified inception module (small scales) decreases RMSE by about 4 points indicating the importance of wider networks in first layers as the critical layers of the network for small object detection.
	Replacing the baseline's extra layers with more modified inception models further improves the performance.
	This highlights the role of higher-resolution layers in the vehicle detection tasks.
		\begin{table*}
			\centering
			\caption{Evaluation of modified inception module (mincep) in the stage 5 on the CARPK dataset. The DAB units are in place. Smaller the RMSE, better the performance.}
			\begin{adjustbox}{width=0.7\textwidth}
				\label{tab:1}
				\begin{tabular}{c|c||c|c|c|c|c}
					method& RMSE    & small scales & mincep-1 & mincep-2 &  mincep-3 & mincep-4 \\
					\toprule\toprule
					ShuffleNet-SSD-512 &    63.57       &- &- &- & - &\\
					ShuffleDet &          52.75         & - & -& -& -&\\
					ShuffleDet &          45.26         & \checkmark& -& -& -&\\
					ShuffleDet &          41.89       & \checkmark & \checkmark& -& -& -\\
					ShuffleDet &           40.47      & \checkmark & \checkmark& \checkmark&- & -\\
					ShuffleDet &          39.67      &\checkmark & \checkmark& \checkmark& \checkmark& -\\
					ShuffleDet &           38.46     & \checkmark  & \checkmark& \checkmark& \checkmark&\checkmark \\
				\end{tabular}
			\end{adjustbox}
		\end{table*} 

Table~\ref{tab:2} represents the evaluation of DAB unit in which we observe a significant reduction in RMSE (almost 5 points) even by the first DAB unit on stage 2. This further indicates the significance of including higher-resolution layer.
Furthermore, the results show that adding DAB modules to the extra layer can additionally enhance the performance to a lesser degree.
This performance indicates that applying the DAB unit in the high-resolution layers can lead to a significant improvement in detecting small vehicles allowing a better utilization of the deformable convolution to adapt to the vehicle geometries.
	
    \begin{table*}

		\label{tab:2}
		\centering
		\caption{Evaluation of using DAB unit on the CARPK dataset. We refer to modified inception layers as {\fontfamily{qcr}\selectfont mincep}. The modified inception modules and small scales are in place.}
		\begin{adjustbox}{width=0.8\textwidth}
\label{tab:2}
			\begin{tabular}{c|c||c|c|c|c|c|c}
				method&  RMSE  & DAB-stage2 & DAB-stage3 & DAB-stage4 &  DAB-mincep-1 & DAB-mincep-2 & DAB-mincep-3 \\
				\toprule\toprule
				ShuffleNet-SSD-512 &   63.57   &- &- &- & - & -& -\\
				ShuffleDet &          49.26   & - & -& -& - & - & -\\
				ShuffleDet &       44.17       & \checkmark& -& -& -& -& -\\
				ShuffleDet &         42.02     & \checkmark& \checkmark&- & -& -& -\\
				ShuffleDet &          40. 75   & \checkmark& \checkmark& \checkmark& -&- & -\\
				ShuffleDet &            39.81  & \checkmark& \checkmark& \checkmark&\checkmark &- & -\\
			ShuffleDet &            39.14         & \checkmark& \checkmark& \checkmark&\checkmark &\checkmark & -\\
				ShuffleDet &            38.46     & \checkmark& \checkmark& \checkmark&\checkmark&\checkmark&\checkmark
			\end{tabular}
		\end{adjustbox}

	\end{table*} 
	
\begin{figure*}[h]
				\begin{subfigure}{0.47\textwidth}
					\includegraphics[width=\textwidth]{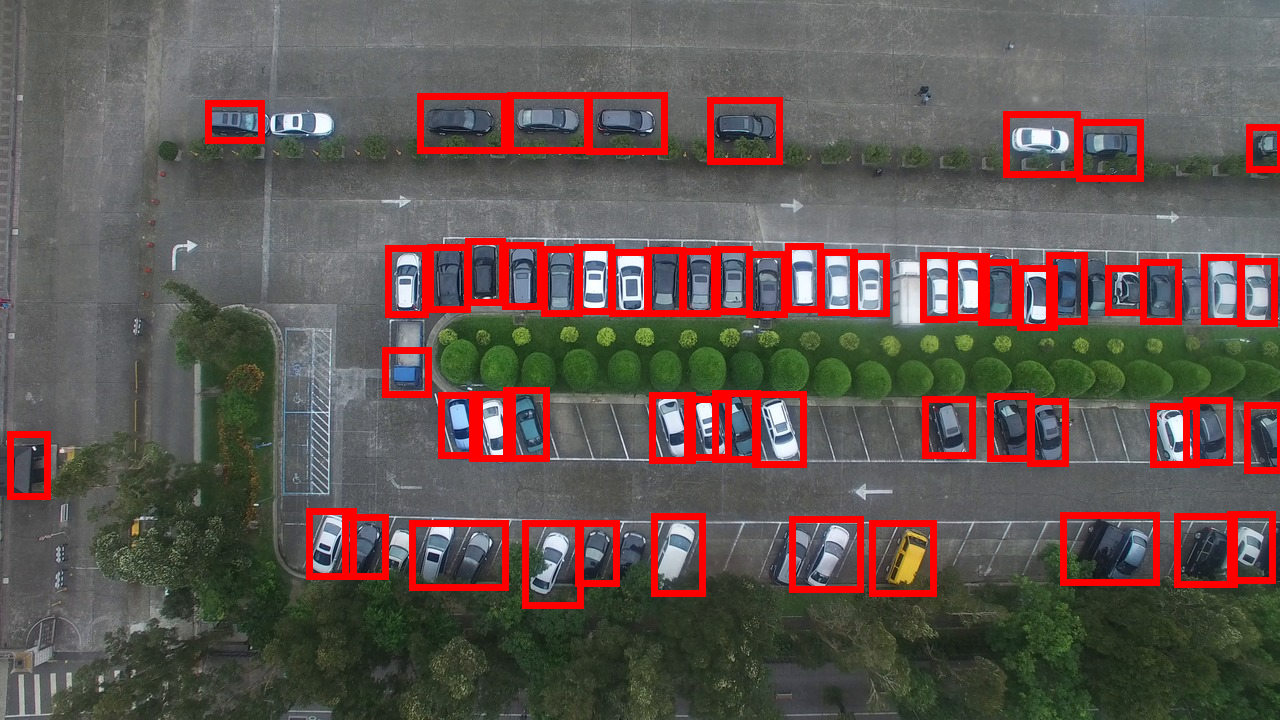}
					\caption{}
					\label{}
				\end{subfigure}
				 \hfill
				\begin{subfigure}{0.47\textwidth}
					\includegraphics[width=\textwidth]{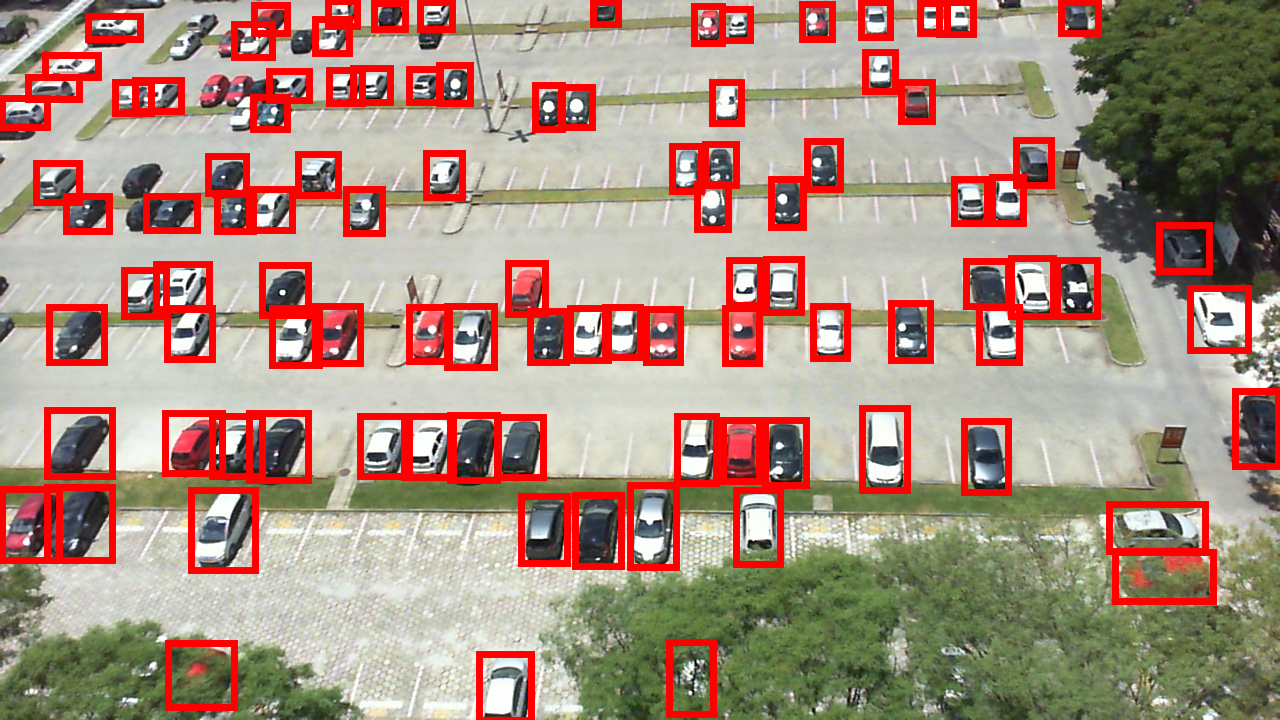}
					\caption{}
					\label{}
				\end{subfigure}
				\caption{Sample vehicle detection results using ShuffleDet on the CARPK(a) dataset and the PUCPR+ dataset(b).}
				\label{fig:carpk}

			\end{figure*}

We choose $s_{min}=0.05$ and $s_{max}=0.4$ as minimum and maximum vehicle scales with ratio of ${2,3,1/2,1/3}$ as hyper-parameters in the original SSD.
This improves the performance significantly according to Table~\ref{tab:1} by almost 7 RMSE points.
It is worth noting that ShuffleNet-SSD-512 has 2.94 GFLOPs as complexity cost while ShuffleDet has 3.8 GFLOPs.
This shows ShuffleDet adds only a marginal computation cost while achieving a significant boost in the accuracy.
Figure~\ref{fig:carpk} shows sample results of ShuffleDet on the CARPK and PUCPR+ datasets.
	\subsection{Comparison with the benchmark}
In this part, compare our method with the benchmark.
Tables~\ref{tab:3}~and~\ref{tab:4} show that our method can achieve competitive performance while having significantly less computation cost compared with the state of the art.
In comparison with the original implementation of Faster-RCNN~\cite{fasterrcnnNIPS2015} and Yolo~\cite{yolov1}, our method achieves significantly better results.
ShuffleDet achieves comparative result with the state of the art with only about less 2 RMSE points in the CARPK dataset.
The reason for the big gap between SSD-512, MobileNet-SSD-512 and shuffleDet is mostly due to our tuned scales and aspect ratios.
This effect can also be observed between the original implementation of Faster-RCNN with and without small RPNs.
\begin{table*}
	\label{tab:3}
	\centering
	\caption{Evaluation of ShuffleDet with the benchmark on the PUCPR+ dataset. The less is better.}
	\begin{adjustbox}{width=0.75\textwidth}
		\label{tab:3}
		\begin{tabular}{c||c|c|c|c}
			method& backbone & GFLOPs & MAE & RMSE \\
			\toprule\toprule
			YOLO\cite{yolov1} & custom & 26.49 & 156.00 & 200.42  \\
			Faster-RCNN\cite{fasterrcnnNIPS2015} & VGG16 & 118.61 & 111.40 & 149.35\\
			Faster R-CNN (RPN-small)\cite{fasterrcnnNIPS2015} & VGG16 & 118.61 & 39.88 & 47.67\\
			One-Look Regression\cite{DBLP:journals/corr/MundhenkKSB16} & - & - & 21.88 & 36.73\\
			Hsieh et al.\cite{HsiehLH17} & VGG16 & - & 22.76 & 34.46\\
			SSD-512\cite{DBLP:conf/eccv/LiuAESRFB16}& VGG16 & 88.16 & 123.75& 168.24\\
			MobileNet-SSD-512\cite{8099834}& MobileNet & 3.2 & 175.26 & 225.12\\		
			our ShuffleDet & ShuffleNet & 3.8  & 41.58 & 49.68\\
		\end{tabular}
	\end{adjustbox}
\end{table*} 
\begin{table*}
	\centering
	\caption{Evaluation of ShuffleDet with the benchmark on the CARPK dataset. The less is better.}
	\begin{adjustbox}{width=0.75\textwidth}
		\label{tab:4}
		\begin{tabular}{c||c|c|c|c}
			method& backbone & GFLOPs & MAE & RMSE \\
			\toprule\toprule
			YOLO\cite{yolov1} & custom & 26.49 & 48.89 & 57.55  \\
			Faster-RCNN\cite{fasterrcnnNIPS2015} & VGG16 & 118.61 & 47.45 & 57.39\\
			Faster R-CNN (RPN-small)\cite{fasterrcnnNIPS2015} & VGG16 & 118.61 & 24.32 & 37.62\\
			One-Look Regression\cite{DBLP:journals/corr/MundhenkKSB16} & - & - & 59.46 & 66.84\\
			Hsieh et al.\cite{HsiehLH17} & VGG16 & - & 23.80 & 36.79\\
			SSD-512\cite{DBLP:conf/eccv/LiuAESRFB16}& VGG16 & 88.16 & 48.02 & 57.42\\
			MobileNet-SSD-512\cite{8099834}& MobileNet & 3.2 & 57.34 & 65.24\\		
			our ShuffleDet & ShuffleNet & 3.8  & 26.75 & 38.46\\
		\end{tabular}
	\end{adjustbox}
\end{table*}

Moreover, ShufflDet achieves its superiority to Faster-RCNN and Yolo while it is significantly more computation efficient, 3.8 GFLOPs compared to 118 and 26.49 GFLOPs.
While Faster-RCNN runs at Jetson TX2 with 1 FPS, tiny Yolov2 at 8 and Yolov2 at 4 FPS, and original SSD with 88.16 GFLOPs at 5 FPS, our ShuffleDet network runs at 14 FPS showing a great potential to be deployed in the real-time on-board processing in UAV imagery.
In addition, our approach achieves almost 70\% and 50\% better performance than MobileNet-SSD-512 and the naive implementation of ShuffleNet-SSD on the CARPK dataset, relatively.

	\section{Generalization Ability}
\begin{figure*}[h]
			\centering
			\includegraphics[width=0.7\textwidth]{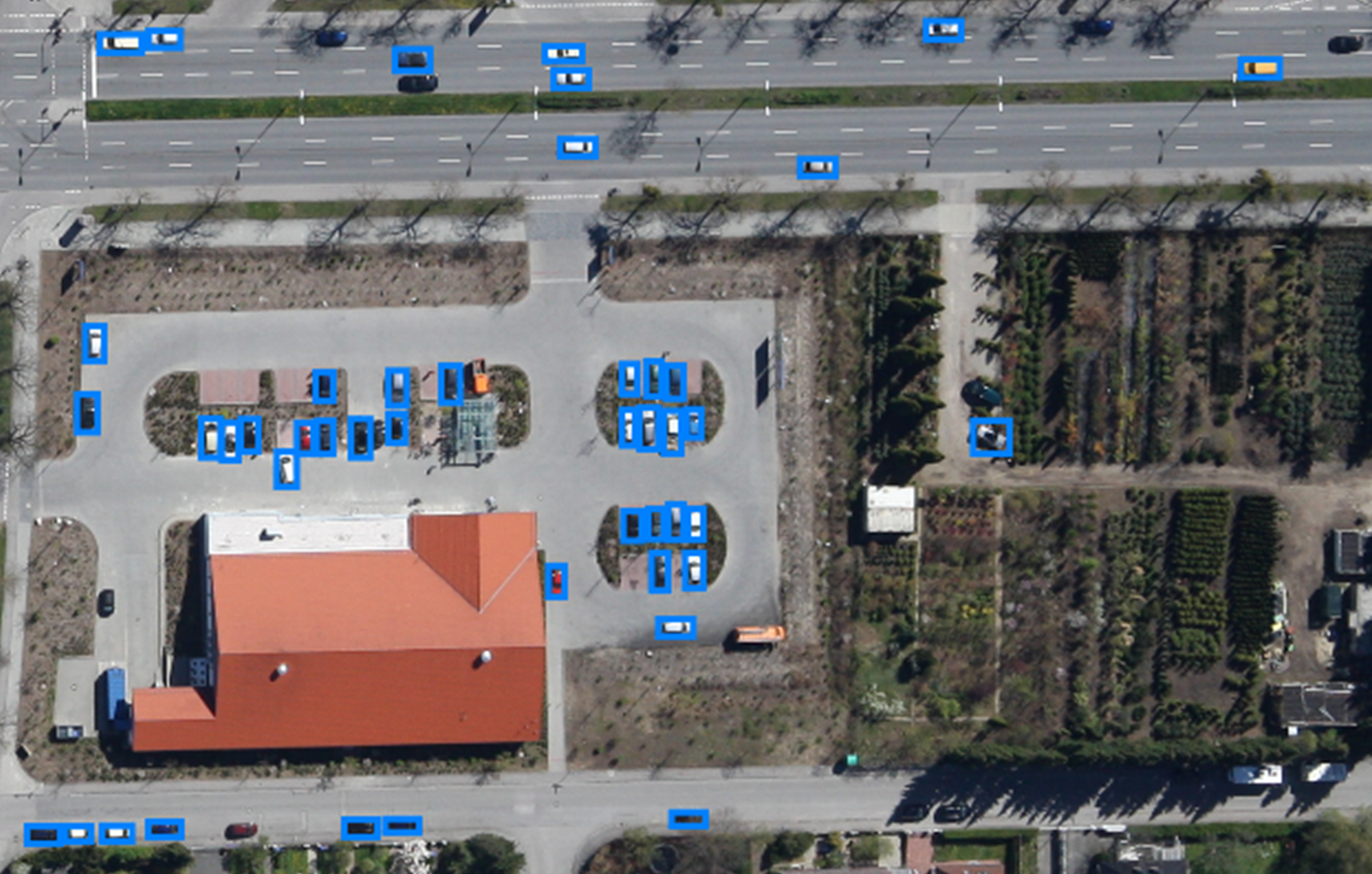}
			\caption{Vehicle detection result using ShuffleDet on the 3K-DLR-Munich dataset.}
			\label{fig:dlr}
		\end{figure*}

To evaluate the generalization ability of our method, we train it on the 3K-DLR-Munich dataset~\cite{KangMattyus}.
This dataset contains aerial images of $5616\times3744$ pixels over the Munich city.
Due to the large size of each image similar to~\cite{azimiACCV}, we chop the images into the patches of $512\times512$ pixels which have 100 pixels overlap. To prepare the final results, for each image, we merge the detections results of the patches and then apply none-maximum suppression. 
Figure~\ref{fig:dlr} illustrates a detection result of our algorithm for the 3K-DLR-Munich dataset.
        
Table~\ref{tab:dlr} compares the performance of ShuffleDet and two implementations of Faster-RCNN on the 3K-DLR-Munich dataset. According to the table, ShuffleDet not only outperforms the Faster-RCNN methods but also its inference is much more time efficient.
The consistent behavior of our proposed approach on the 3K-DLR-Munich dataset indicates that it could be generally applied to different datasets. 
ShuffleDet is capable of 2 FPS processing of high-resolution aerial images in Jetson TX2 platform while Faster-RCNN with VGG16 and ResNet-50 takes a couple of seconds.

	\begin{table*}
		\centering
		\caption{Evaluation of ShuffleDet on 3K-DLR-Munich dataset. Inference time is computed in Jetson TX2 as an edge device.}
		\begin{adjustbox}{width=0.75\textwidth}
			\label{tab:dlr}
			\begin{tabular}{c||c|c|c|c}
				method& backend & GFLOPs & mAP & inference time\\
				\toprule\toprule
				Faster-RCNN~\cite{fasterrcnnNIPS2015} & VGG-16 & 118.61 & 67.45\% & 7.78s \\
				Faster-RCNN~\cite{fasterrcnnNIPS2015} & ResNet-50 & 22.06 &69.23\% & 7.34s  \\
				our ShuffleDet & ShuffleNet & \textbf{3.8} & 62.89 & \textbf{524ms}\\
			\end{tabular}
		\end{adjustbox}
	\end{table*}
    
	\section{Conclusions}
In this paper, we presented ShuffleDet, a real-time vehicle detection algorithm appropriate for on-board embedded UAV imagery.
ShuffleDet is based on channel shuffling and grouped convolution in its feature extraction stage.
To evaluate the effect of different modules of ShuffleDet, an ablation study is performed to discuss its accuracy and time-efficiency.
Joint channel shuffling and grouped convolution significantly boost the inference time.
Inception modules with depthwise convolutions enhance the accuracy while introducing a marginal computation burden.
Moreover, we show residual modules with deformable convolutions are effective modules for semantic representation enhancement in the small number of layers as well as domain adaptation.
Experimental results on the CARPK and PUCPR+ datasets indicate that ShuffleDet outperforms the state-of-the-arts methods while it is much more time and computation efficient. Additionally, the consistent behavior of ShuffleDet on the 3K-DLR-Munich dataset demonstrate its generalization ability.
Furthermore, the implementation of ShuffleDet on Jetson TX2, which runs at 14 FPS, showing a great potential of our approach to be used in UAVs for on-board real-time vehicle detection.
\renewcommand{\bibname}{References}
\bibliographystyle{splncs04}
\bibliography{egbib}
\end{document}